\documentclass[journal]{IEEEtran}
\usepackage{cite}
\usepackage{amsmath,amssymb,amsfonts}
\usepackage{algorithmic, algorithm}
\usepackage{graphicx,color}
\usepackage{textcomp}
\usepackage{comment}
\usepackage[hidelinks]{hyperref}
\usepackage{cleveref}
\usepackage{multirow}
\usepackage[caption=false,font=footnotesize]{subfig}
\usepackage[dvipsnames]{xcolor}

\def\BibTeX{{\rm B\kern-.05em{\sc i\kern-.025em b}\kern-.08em
    T\kern-.1667em\lower.7ex\hbox{E}\kern-.125emX}}
\begin{document}

\crefname{figure}{Fig.}{Figs.}
\crefname{table}{Table}{Tables}
\crefname{algorithm}{Algorithm}{Algorithms}
\crefname{section}{Section}{Sections}

\title{Tram-FL: Reducing Communication and Computation Costs through Sequential Model Circulation in Decentralized Federated Learning}

\author{Kota~Maejima,
        Takayuki~Nishio,
        Asato~Yamazaki,
        and~Yuko~Hara-Azumi%
\thanks{This work has been submitted to the IEEE for possible publication. Copyright may be transferred without notice, after which this version may no longer be accessible.}%
\thanks{K. Maejima, T. Nishio, A. Yamazaki, and Y. Hara-Azumi are with the School of Engineering, Institute of Science Tokyo, Tokyo 152-8550, Japan. Corresponding author: Takayuki Nishio (e-mail: nishio@ict.eng.isct.ac.jp).}%
\thanks{This work was supported in part by JST PRESTO Grant Number JPMJPR2035, ASPIRE Grant Number JPMJAP2346, and JSPS KAKENHI Grant Number JP23H00464.}}

\markboth{}{Maejima \MakeLowercase{\textit{et al.}}: Tram-FL: Sequential Model Circulation for Decentralized Federated Learning}

\maketitle

\begin{abstract}\label{s:abstract}
Conventional decentralized federated learning (DFL) often focuses on clients, with each client maintaining a model copy, performing updates individually, and undertaking model exchange and integration. While fully leveraging computational resources can shorten training times, it can also lead to significant computational and communication waste. This is especially pronounced with non-independent and identically distributed (non-IID) data, where achieving high model accuracy demands extra resources. This research shifts focus to the model itself, aiming to realize DFL with minimal computation and communication costs. To this end, we propose Tram-FL (Traveling Model Training Mechanism for Decentralized Federated Learning), a mechanism designed to efficiently address these challenges. It sequentially trains a single model by circulating it among nodes. We address the training scheduling problem in model circulation-based training, specifically determining which nodes should update the model and the number of updates to perform. This is approached by considering the model’s circulation route and update iteration allocation, for which we propose simple yet effective methods. Additionally, with quantized momentum, Tram-FL achieves high accuracy with fewer model circulations while controlling communication load per transmission. Experimental results show that the proposed algorithm, even with non-IID data, converges to a global model with reduced communication and computation.

\end{abstract}


\section{INTRODUCTION}\label{s:introduction}
\IEEEPARstart{T}{HE} extensive data aggregation for machine learning poses a severe threat to privacy \cite{ren2018querying}. This concern motivates the adoption of federated learning (FL) \cite{mcmahan2017communication}, where participants can collaboratively learn a model by synchronizing only the locally trained model parameters without revealing their raw data.
In conventional federated learning, a central parameter server coordinates a large consortium of participating clients, aggregating gradients or model parameters from clients to collaboratively advance the learning of a global model \cite{li2020federated_learn}.
While this FL system is promising, they face numerous challenges. In some instances, large organizations can serve the role of a central server in internet of things (IoT) applications, yet in many FL environments, finding a reliable and robust central server is challenging. Furthermore, the failure of the central server introduces a single point of failure for the entire network \cite{kairouz2021advances}.

Hence, a decentralized federated learning (DFL) framework has been proposed. This framework eliminates the need for a central server and synchronizes FL updates and aggregation among data nodes. Gossip averaging \cite{boyd2006randomized, dimakis2008geographic, khosravi2016broadcast} is a well-known method in diverse, decentralized algorithms, allowing different nodes within a network to exchange information in a peer-to-peer manner without the aid of a central server. \cite{blot2016gossip, Jin2016HowTS, ormandi2013gossip, koloskova2019decentralized, koloskova2020unified} shows that gossip averaging, combined with stochastic gradient descent (SGD), is used to train deep learning models in a decentralized manner, exhibiting superior convergence properties. Building on these studies, gossip averaging was subsequently applied to the DFL scheme, where model updates and averaging are alternately conducted at local nodes \cite{hegedHus2019gossip, roy2019braintorrent, savazzi2020federated, hu2019decentralized}. This averaging consensus technique is employed in DFL and is commonly used in Consensus-Based distributed SGD (CDSGD) \cite{jiang2017collaborative}, decentralized parallel SGD (D-PSGD) \cite{lian2017can}, and related algorithm \cite{Tang2018D2DT}.

\begin{figure}[!t]
    \centering
    \begin{tabular}{cc}
    \subfloat[Conventional DFL]{\includegraphics[width=0.45\columnwidth]{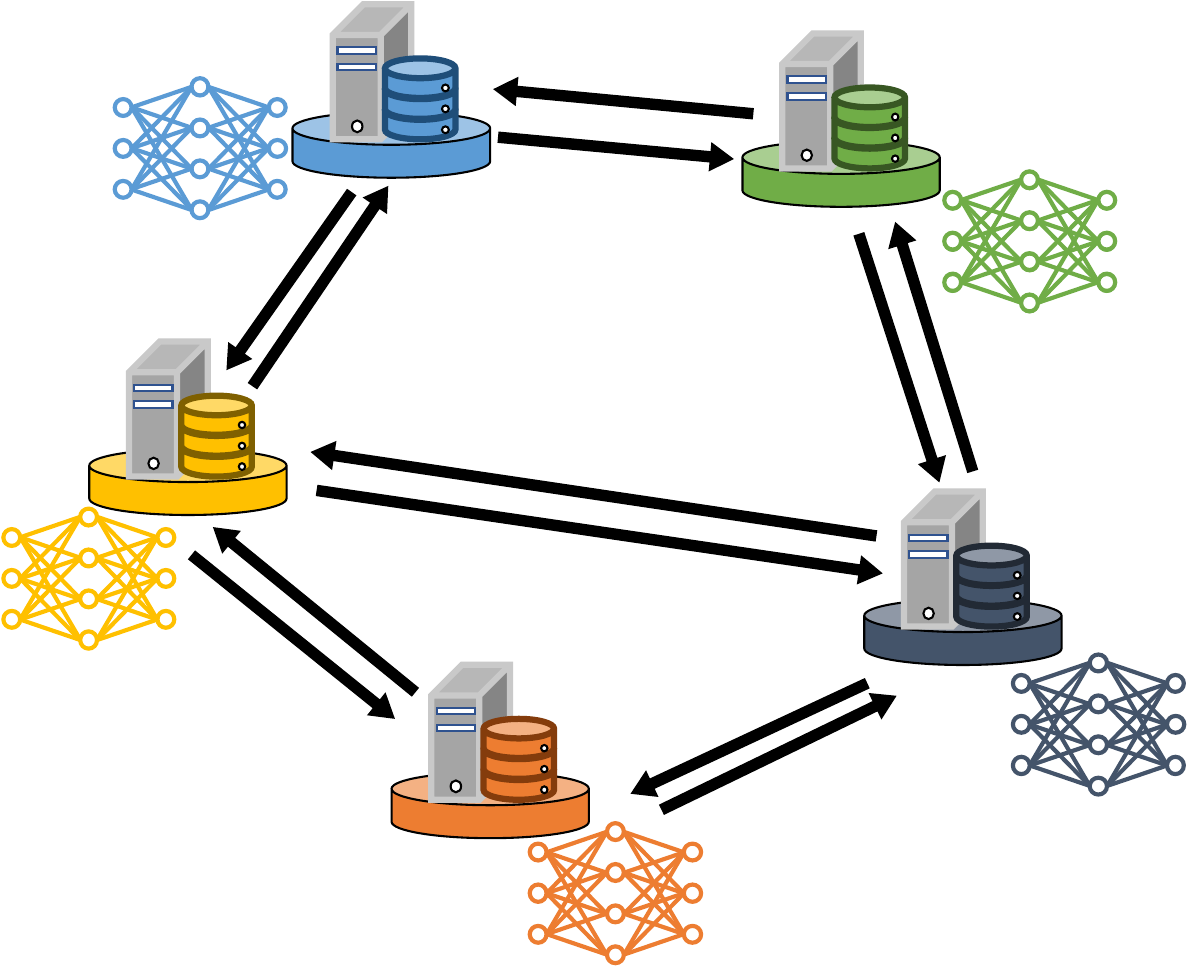}} &
    \subfloat[Ours: Tram-FL]{\includegraphics[width=0.45\columnwidth]{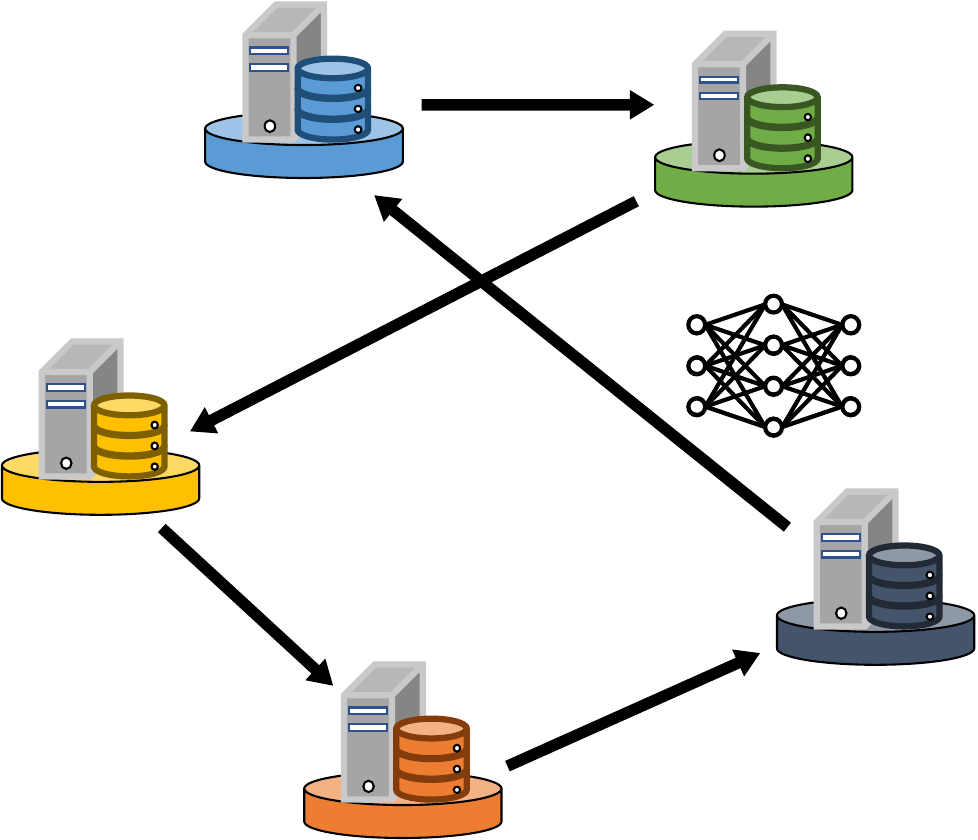}} \\
    \end{tabular}
    \caption{Training processes of the conventional DFL methodology alongside our proposed approach Tram-FL.}
    \label{fig:compare_process}
\end{figure}

The previously mentioned DFL algorithms surmount several challenges inherent to conventional FL systems that necessitate a central server, yet they are not invariably efficient. \cref{fig:compare_process} illustrates the training processes of the conventional DFL methodology alongside our proposed approach. As depicted in \cref{fig:compare_process}\,(a),  in conventional DFL, a multiplicity of learning models exist within the network, advancing in proficiency through frequent inter-nodal communication and updates. Consequently, the communication costs between clients and the computational costs of the clients often become significant bottlenecks. Furthermore, since different nodes possess distinct training datasets, usually non-overlapping and with diverse distributions (known as non-IID data), the optimal points of each node's loss function can be substantially divergent, potentially far from the global optimum \cite{ZHU2021371}. Consequently, the aggregated model may deviate substantially from the global optimum, necessitating additional communication and computation to achieve a high-precision global model.

Numerous studies have been conducted to address this issue. Methods have been proposed that impose constraints on model updates \cite{niwa2020edge}, apply techniques such as mutual knowledge transfer (MKT), quadratic programming (QP), and knowledge distillation \cite{he2020group, esfandiari2021cross, aketi2024cross}, utilize the similarity of label distributions \cite{onoszko2021decentralized, bellet2022d}, and effectively control updates and communications \cite{liao2023adaptive, yamazaki2023convergence}. However, while these methods improve the convergence of models in non-IID settings, they still involve frequent updates and communications across multiple models, thus resulting in substantial computational and communication costs. Furthermore, although their effectiveness in non-IID settings with minor data skewness has been demonstrated in scenarios with significant data skewness, achieving a high-accuracy global model still necessitates additional communication and computation.

To address the aforementioned challenges, we introduce a novel DFL framework named Tram-FL. Unlike existing methodologies, Tram-FL enhances learning by circulating a single global model among nodes as \cref{fig:compare_process}\,(b). In other words, instead of utilizing local models or model aggregation, each node directly updates the global model. 
By carefully scheduling the circulation-based model training, specifically determining the nodes through which the model circulates and the number of update iterations, it is feasible to mitigate the accuracy degradation caused by non-IID data. For example, if certain nodes contain data primarily from Classes 1-3, 4-6, and 7-9, respectively, circulating the model sequentially through these nodes helps prevent the model from becoming biased toward any specific class.
Furthermore, compared to traditional DFL, Tram-FL achieves quicker model convergence and requires only a single model to be learned within the network, thereby significantly reducing both computational and communication costs.

The scheduling problem is complex, as it must consider network topology, distribution of data held by each node, and their computational capabilities. 
This paper proposes an approach where, under a simplified scenario, the number of update iterations and the determination of routes are addressed disjointedly, offering a streamlined solution to this complex problem. We define the method of determining the update iterations number as `mini-batch number allocation' and the decision of a suitable circulation route as `route determination.' The proposed mini-batch number allocation determines the number of update iterations by minimizing the distribution bias in the data used for previous model training. Furthermore, the proposed route determination selects a route that avoids short-term data biases while imposing constraints of consecutive training on the last node of the cycle.

We further augment the aforementioned methodology by adding the quantized momentum. Specifically, during model training, momentum is computed and transmitted to the next node along with the model parameters. However, transmitting momentum as-is would increase communication costs, leading to a trade-off between model accuracy and communication expenses. Therefore, we adopt a technique of compressing and transmitting the momentum under the assumption that the precision of momentum does not significantly impact training.

To evaluate the efficacy of Tram-FL, we conducted experiments utilizing multiple datasets. These experiments involved a comparative analysis of model accuracy per communication traffic and model update with existing methodologies. Furthermore, we assessed the impact of momentum compression on model accuracy. The results demonstrated that Tram-FL achieved stable convergence with significantly reduced communication traffic and model updates. Additionally, it was evidenced that compression of momentum up to a certain threshold offers greater benefits in reducing communication traffic than the corresponding deterioration in accuracy.

The contributions of this paper are summarized as follows:
\begin{itemize}
    \item We introduce Tram-FL, a novel approach that enhances learning by circulating a single global model among nodes. Unlike traditional DFL methods, which frequently employ local models and model aggregation, Tram-FL operates by allowing each node to directly update the global model. This methodology significantly reduces communication and computational costs by eliminating the need for frequent updates and transmissions of multiple models, a prevalent bottleneck in conventional DFL techniques.
    \item We introduce mini-batch number allocation to determine the optimal number of update iterations for each node, route determination to decide the most efficient circulation path for the model, and the momentum compression and transmission. These enhancements augment the learning efficiency of Tram-FL and achieve further reductions in communication and computational expenses.
\end{itemize}

This paper extends our previous work \cite{maejima2023tram}, which provided a preliminary demonstration of a model circulation-based DFL mechanism. In this paper, we further advance the initial framework by developing a method for adjusting the number of mini-batches used in each node's local training, along with the quantized momentum, leading to significant alterations in the method of determining the model's circulation route to ensure compatibility with the proposed mini-batch number allocation. Additionally, we performed extensive evaluations to thoroughly assess the effectiveness of these improvements.

The remainder of this manuscript is structured as follows:
In the \cref{s:related_works}, we describe conventional DFL and research aimed at improving model accuracy in non-IID settings.
In the \cref{s:proposed_method}, we elaborate on the system model, problem formulation, the mini-batch number allocation, the method for determining cyclical routes, and the approach for momentum compression.
In the \cref{s:evaluation}, we assess the proposed method through experiments utilizing an image classification dataset.
In the \cref{s:conclusion}, we articulate the conclusions of this manuscript and propose directions for future research.

\section{RELATED WORKS}\label{s:related_works}
As delineated in the \cref{s:introduction}, within non-IID settings, DFL encounters formidable challenges in achieving a high-accuracy global model. Numerous studies have been conducted to address this issue.

PDMM-SGD algorithm \cite{niwa2020edge} integrates the PDMM algorithm \cite{sherson2018derivation, zhang2017distributed}, originally designed for convex model optimization on P2P networks, into deep neural network (DNN) cost minimization problems, thereby reformulating the constrained minimization problem. The reconstructed problem incorporates linear consensus constraints, which advocate for uniformity of model variables across nodes in the network. This constraint ensures that the optimal point of each node's loss function coincides with the global optimum.
\cite{he2020group} effectively incorporates the benefits of MKT into the DFL framework. In each round of the proposed Def-KT algorithm, local models are transmitted to a randomly selected node. In contrast to averaging, each recipient node fuses the local model with the received model using MKT. Through MKT, two models with distinct knowledge learn from each other, resulting in a model that indirectly acquires knowledge from both sources.
\cite{esfandiari2021cross} introduces a cross-gradient aggregation method to address the challenges posed by non-IID data. This technique entails collecting derivatives of a model with respect to the datasets of neighboring nodes. The gathered gradients are then aggregated and utilized to update the model through projected gradients, which are optimized via quadratic programming (QP).
\cite{aketi2024cross} is a methodology derived from applying knowledge distillation to DFL, introducing two additional contrastive loss terms related to cross-features in each node. Cross-features encompass feature quantities obtained by inputting local data into the model of adjacent nodes, as well as those acquired by feeding data from neighboring nodes into the local model. Minimizing the contrastive loss function enhances the similarity among local models, thereby improving model performance and coherence.

There are attempts to improve the convergence of the data-heterogeneous DFL by controlling the topology of peer-to-peer networks. This involves selectively choosing peers for the exchange of models and gradients.
\cite{onoszko2021decentralized} enables nodes with similar data distributions to detect each other, collaboratively evaluate training losses on each other's data, and learn models suited to their local data distributions.
\cite{bellet2022d} proposes a novel topology for DFL, grouping nodes into sparsely connected cliques. This strategy ensures that the label distribution within cliques represents the global label distribution, thereby reducing gradient bias.
\cite{liao2023adaptive} integrates adaptive control of both local update frequency and network topology. It establishes a theoretical relationship between local update frequency and network topology in terms of model training performance and seeks an upper bound on convergence. Based on this, it proposes an optimization algorithm that adaptively determines local update frequency and constructs network topology.
\cite{yamazaki2023convergence} facilitates the reflection of data from all nodes on all models in the network, even in settings with large network topology bias, by efficiently using both swap, which exchanges models, and skip, which acts like a hub and sends one node's model to another node. 

These methodologies have addressed several challenges faced by DFL in non-IID settings. However, all of them retain the premise of maintaining multiple models in parallel across the network. Under this premise, two costs remain structural: the training computation, which is proportional to the number of models maintained in the network, and the recurrent model exchanges required for aggregation or consensus. \cref{tb:overhead_comp} summarizes this point analytically: with $n$ nodes of average degree $d$ and model size $M$, gossip- and consensus-based methods transmit $O(nd)$ model copies per round and train $n$ models in parallel, whereas circulating a single model reduces the transmissions to $n$ transfers of a single model (plus compressed momentum $M_\mathrm{mom}$) per cycle---i.e., from $O(nd)$ to $O(n)$ model transmissions per round---and the number of maintained models to one. Reducing the number of models in the network to one is the design point that eliminates both structural costs by construction, which partial refinements of parallel DFL cannot reach. This observation motivates the model-circulation approach proposed in this paper, targeting accurate global models in highly skewed non-IID settings with minimal communication and computational costs.

\begin{table}[t]
  \caption{Analytical comparison of communication and computation per training round, where one round lets every node update once (for Tram-FL, one cycle). $n$: nodes, $d$: average degree, $M$: model size, $M_\mathrm{mom}$: compressed momentum size.}
  \label{tb:overhead_comp}
  \centering
  \begin{tabular}{l|cc}
  \hline
    & Transmissions per round & Models trained \\ \hline
    Gossip-SGD & $O(nd) \times M$ & $n$ \\
    PDMM-SGD & $O(nd) \times M$ & $n$ \\
    Tram-FL & $n \times (M + M_\mathrm{mom})$ & $1$ \\ \hline
  \end{tabular}
\end{table}

\section{PROPOSED METHOD}\label{s:proposed_method}
This section presents the proposed Tram-FL. In a nutshell, Tram-FL trains a single global model by circulating it among nodes: at the beginning of each cycle, the number of training mini-batches for every node is allocated so as to balance the cumulative label usage (Step~1), the circulation route is determined accordingly (Step~2), and the model and momentum are then updated at each node (Step~3) and forwarded with quantized momentum to the next node (Step~4), as illustrated in \cref{fig:iteration_process} and summarized in \cref{alg:Tram-FL}. In the following, \cref{ss:system_model} describes the system model, \cref{ss:algorithm} details the training mechanism, \cref{ss:num_of_batch_alg} and \cref{ss:model_routing_alg} present the mini-batch number allocation and the route determination, respectively, \cref{ss:momentum_comp} describes the momentum compression, and \cref{ss:summary} summarizes the overall procedure.

\subsection{System Model}\label{ss:system_model}
FL can be categorized into two distinct types: cross-device FL and cross-silo FL \cite{kholod2020open}. In cross-device FL, nodes are typically small-scale distributed entities, such as smartphones, wearables, and edge devices. These nodes possess only a modest amount of local data. Participation in each training round is restricted to a mere fraction of these nodes. Hence, the success of cross-device FL typically necessitates the involvement of a considerable number of edge devices, potentially reaching up to several million. Conversely, in cross-silo FL, nodes are customarily entities such as corporations or organizations (like hospitals or banks). The number of participants is relatively limited (ranging from 2 to 100 entities), and each node is expected to participate in the entire training process. This paper presupposes that the data servers involved in cross-silo FL (hereinafter referred to as 'nodes') collaboratively learn a DNN model for a specific shared application.

In this study, we assume that corporate data servers in cross-silo FL act as nodes. Each node possesses sufficient computational power to train a model on its local data within a reasonable time frame, and the nodes communicate with each other over the Internet with acceptable latency; we assume reliable logical links with no network disconnections during training. The communication traffic generated when exchanging models between nodes is treated as the communication overhead of the training process, and our proposed method aims to improve the trade-off between this communication traffic and the accuracy of the trained model. Issues related to reducing communication and computation latency for greater real-time performance, which require considering heterogeneous communication bandwidths and computation power among nodes, are beyond the scope of this paper and will be explored in future work.

We note, however, that the circulation mechanism itself offers a natural recovery path against node failures: because the entire training state is carried by the circulating model and momentum, a failure is detected by a transmission timeout; since the preceding node retains its copy of the model and momentum until the transfer is acknowledged, it can re-route the model to the next reachable node in the order of the allocated batch counts, and the failed node's allocation can be re-distributed in the next cycle by re-solving the allocation with the node excluded. The cost of such a failure is thus the loss of at most one node's worth of updates, not the loss of the training state. A quantitative evaluation of robustness under node churn is left as future work.

Because the model visits nodes sequentially, the wall-clock time of one cycle grows linearly with the number of nodes, and a single circulating model cannot exploit parallel computation across nodes. Tram-FL is therefore designed for cross-silo FL, where the number of participants typically ranges from several to around one hundred~\cite{kholod2020open}, rather than for cross-device FL with thousands of devices. For larger networks, a natural extension is to partition nodes into clusters and circulate one model per cluster with periodic inter-cluster synchronization, trading a part of the communication savings for parallelism; we leave this hierarchical extension to future work.

This paper assumes %
each node holds data that has been subjected to proper preprocessing steps, such as labeling and imputation of missing values, and it is postulated that the nodes are capable of training models using their own data. For the purpose of model training, nodes construct mini-batches from their data through random sampling. This assumption is ubiquitous in the field of FL. Particularly in a cross-silo setting, where each node represents an enterprise or organization, this assumption is commonplace due to the high reliability of the data.

Furthermore, it is anticipated that statistical information about the datasets owned by the nodes, such as label distributions, can be shared in advance. Specifically, the distribution of data labels across all nodes participating in the training is shared beforehand. This information is crucial in determining the number of training batches and the route of the model, and it is a common assumption in non-IID settings of FL setups \cite{Duan2019, luo2022fedsld}.

We also note the privacy implication of this assumption. What is shared is only the per-label sample counts $L_i$, i.e., a $|C|$-dimensional histogram, rather than raw data, features, or gradients; the risk of reconstructing individual samples from such coarse statistics is considerably lower than that of gradient-sharing schemes, which are known to be vulnerable to gradient inversion attacks~\cite{zhu2019dlg}. Nevertheless, label histograms may still reveal business-sensitive information (e.g., the case mix of a hospital). In such cases, the histograms can be perturbed by differential-privacy mechanisms or aggregated through secure multi-party computation before being shared; since the proposed allocation depends on $L_i$ only through the expected label counts, it remains compatible with noisy statistics at the cost of a looser approximation of the uniform-distribution objective. A quantitative study of this privacy--utility trade-off is left for future work.
\subsection{Model Training Mechanism of Tram-FL}\label{ss:algorithm}
\begin{figure}[!t]
    \centering
    \includegraphics[width=1\columnwidth]{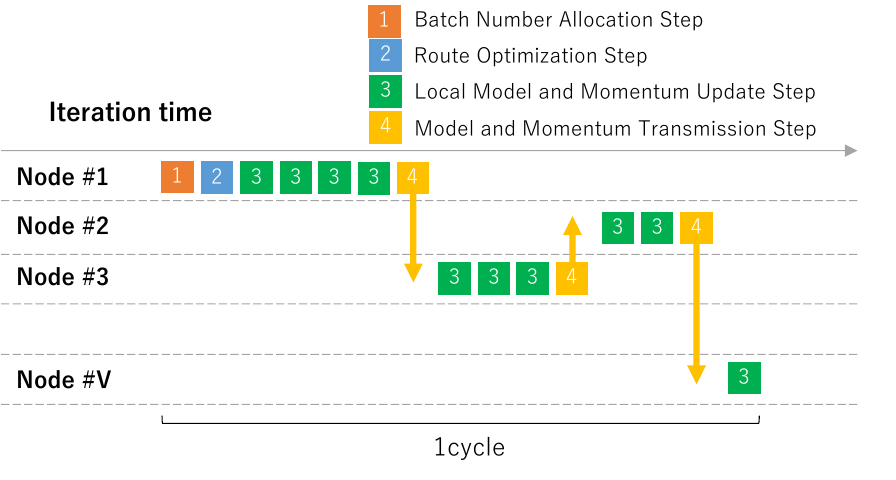}
    \caption{Sequence diagram of the proposed Tram-FL model update procedure. In Step 1, the number of training batches for each node is determined. Subsequently, in Step 2, the model's circulatory route is established, followed by Step 3, where the node updates the model parameters and momentum using mini-batches. Finally, in Step 4, the model and momentum are transmitted to the next node.}
    \label{fig:iteration_process}
\end{figure}

Tram-FL aims to achieve model convergence with minimal communication and computational costs in highly skewed non-IID settings. A pivotal concept in realizing this goal is the circulation of a single model across the network with an optimized number of batches and routes during the learning process.
\cref{fig:iteration_process} illustrates the sequence of model updates and transfers among the nodes. These procedures are delineated into four distinct steps: Step 1 is the mini-batch number allocation step, Step 2 is the route determination step, Step 3 is the local model and momentum update step, and Step 4 is the model and momentum transmission step. Here, one cycle is defined as the global model traversing all nodes participating in the training. At the onset of each cycle, the node that first acquires the global model initiates Step 1: mini-batch number allocation step to determine the number of batches for model updating to be utilized by each node throughout the cycle. Given that the batch size is fixed, this procedure enables the regulation of the update frequency for each node. Subsequently, Step 2: route determination step determines the circulation route of the model for the cycle. Afterward, in Step 3: local model and momentum update step, the model and momentum are updated based on the predetermined batch count. Upon completion of the updates, the model and momentum are transmitted to the subsequent node following the prescribed circulation route in Step 4: model and momentum transmission step. This procedure is iteratively executed until the model validation loss or accuracy reaches a predetermined threshold or until a specified number of communication cycles are completed.

The implementation of Tram-FL is depicted in \cref{alg:Tram-FL}. Let $w^t$ denote the DNN model variables, $v^t$ the momentum variables, and $v_{q}^{t}$ the compressed momentum at model update iteration $t$. The DNN cost function to be minimized is represented as $F(\cdot)$, with momentum compression and decompression denoted by $C(\cdot)$ and $D(\cdot)$, respectively. $\mathcal{V}$ represents the set of nodes, $\alpha$ the momentum coefficient and $\eta$ the learning rate. During $K$ cycles of iteration, Steps 1 through 4 are iteratively executed. 
At the beginning of each cycle, the mini-batch number allocation step determines the batch count for each node to update the model. Following this, the route determination step establishes the model’s circulation route for the cycle. If not the initial model update, decompression of the received compressed momentum from the previous node takes place. The nodes follow the order determined in the route determination step to execute the local model and momentum update step and the model and momentum transmission step. In the local model and momentum update step, the node holding the model updates $w$ and $v$ by extracting the batch count determined in the mini-batch number allocation step from its own training data. During the model and momentum transmission step, the momentum is compressed and then transmitted along with the model to the next node via the determined route in the route determination step. The methods for determining the training batch count for each node and the model's circulation route in the mini-batch number allocation and route determination steps, as well as the method for compressing the momentum, are elaborated in \cref{ss:num_of_batch_alg} to \cref{ss:momentum_comp}.

\begin{figure}[t]
\begin{algorithm}[H]
    \caption{Tram-FL}
    \label{alg:Tram-FL}
    \begin{algorithmic}[1]
    \STATE Initialization of $w^{(0)}$ and $v^{(0)}$
    \STATE $t \leftarrow 0$
    \STATE Locate $w^{(0)}$ to randomly selected node $i$
    \FOR{each cycle $k = 1, 2, 3,\dots, K$}
        \STATE \textcolor{RedOrange}{$\triangleright$ Step 1: Decide number of training batch $x$}
        \STATE \textcolor{RoyalBlue}{$\triangleright$ Step 2: Decide model's circulation route}
        \FOR{each node $i \in \mathcal{V}$}
            \IF{$t$ is not 0}
                \STATE Decompression momentum
                \STATE $v^{t} \leftarrow D(v_{q}^{t})$
            \ENDIF
            \FOR{training batch $x_i$}
                \STATE \textcolor{Green}{$\triangleright$ Step 3: Update model parameters and momentum}
                \STATE \textcolor{Green}{$v^{t+1} \leftarrow \alpha v^{t} - \eta \nabla F(w^{t})$}
                \STATE \textcolor{Green}{$w^{t+1} \leftarrow w^{t} + v^{t+1}$}
                \STATE $t \leftarrow t+1$
            \ENDFOR
            \STATE Compression momentum
            \STATE $v_{q}^{t+1} \leftarrow C(v^{t+1})$
            \STATE \textcolor{Dandelion}{$\triangleright$ Step 4: Transmit model and momentum to next node}
            \STATE \textcolor{Dandelion}{Transmit $v_{q}^{t+1}$ and $w^{t+1}$ to node $j$}
        \ENDFOR
    \ENDFOR
    \end{algorithmic}
\end{algorithm}
\end{figure}

By transmitting a single model, our proposed framework can significantly reduce overall network traffic, offering an advantage over traditional approaches that involve transmitting multiple models. An inherent drawback of this approach is the serialization of model updates, which can potentially prolong the learning duration. However, the serialization also yields substantial benefits when handling non-IID data, and the reason can be explained as follows. In aggregation-based DFL, each node locally descends toward the minimizer of its own empirical risk, whose optimum can be far from the global optimum under label skew; averaging the resulting divergent parameters does not, in general, descend the global objective, a phenomenon known as client drift~\cite{karimireddy2020scaffold}. In contrast, Tram-FL applies all updates to a single model in sequence. From the model's perspective, training is equivalent to stochastic gradient descent on a single virtual data stream formed by concatenating the mini-batches drawn along the circulation route. Our mini-batch number allocation explicitly shapes this stream: by minimizing the variance of the cumulative label histogram, the empirical label distribution of the stream converges to the uniform distribution as cycles proceed. Consequently, in the medium to long term, the update sequence approximates SGD on label-balanced (IID-like) data, avoiding the parameter divergence and cancellation inherent to parallel aggregation. This view is consistent with recent theoretical results showing that sequential FL enjoys better convergence guarantees than parallel FL on heterogeneous data~\cite{li2023sequentialfl}. Two sources of degradation remain: the short-term label bias within a cycle, and the catastrophic forgetting between consecutive visits to a node; the route determination in \cref{ss:model_routing_alg} is designed to suppress both. In addition, reducing the training batch count increases the frequency of model transmission and network traffic. Thus, optimizing both the training batch count and the model's routing path is of paramount importance, and this study primarily focuses on these two elements.

\subsection{Mini-batch Number Allocation}\label{ss:num_of_batch_alg}

In the mini-batch number allocation step, the number of mini-batches each node uses to update the model is determined. In standard machine learning, a fixed number of mini-batches—typically covering all samples in the training dataset—are used to update the model across multiple epochs. However, when client data are non-IID, using a set number of batches and multiple epochs without accounting for data distribution can lead to a model biased towards certain classes. To address this, our proposed method restricts the number of mini-batches each node uses, allowing the model to be updated in a balanced way as it circulates across nodes with varied data, thereby reducing bias.

We define a cycle as the period of time during which the global model travels through all nodes participating in the training. Let's denote $L_i = \{l_{i,c} \mid c \in C\}$ as the set of the number of training samples stored by node $i$, where $C$ signifies a set of labels and $l_{i, c} \geq 0$ indicates the number of samples with label $c$ present at node $i$. The total number of samples possessed by node $i$ is represented by $N_i$, hence $N_i = \sum_{c \in C} l_{i, c}$. Moreover, let $L^k_i$ represent the number of samples utilized by node $i$ for training the received model during cycle $k$, thus $L^k_i = \{l^k_{i,c} \mid c \in C\}$, with $l^k_{i,c} ,(0 \leq l^k_{i,c} \leq l_{i,c})$ being the count of samples with label $c$ used by node $i$ in cycle $k$.

We posit that the cumulative number of samples used for model training up to cycle $k$ is expressed as $L^k$, yielding
\begin{eqnarray}
L^k &=& \Big\{\sum^{k}_{\kappa=1}\sum_{i \in \mathcal{V}} l^{\kappa}_{i, c} \mid c \in C\Big\},
\end{eqnarray}
where $\kappa$ is the cycle index running over the past cycles. For simplified notation, we use $l^k_c$ to represent $\sum^{k}_{\kappa=1}\sum_{i \in \mathcal{V}}l^{\kappa}_{i, c}$.

Our proposed mini-batch number allocation determines the number of training batches for each node such that $L^k$ adheres to a pattern closely approximating a uniform distribution, specifically aiming to achieve 
\begin{equation}\label{eq:obj}
l^k_c = l^k_{c'} \; (c, c' \in C).
\end{equation}
To accomplish this, the proposed algorithm identifies the number of training batches for each node that is expected to minimize the variance of $L^k$. The selection policy can be formulated as follows:
\begin{eqnarray}\label{eq:opt}
x^{*} &=& \underset{x}{\mathrm{argmin}}  \; \sigma (L^k + \sum_{i \in \mathcal{V}} \frac{Bx_{i}}{N_{i}} L_{i})\\
x &=& \{x_{i} \mid i \in \mathcal{V}, \; x_i \in \mathbb{Z}, \; 0 \leq x_i \leq \frac{N_i}{B} \} \\
\sigma(L^k) &=& \frac{1}{|C|} \sum_{c \in C} (l^k_c - \frac{1}{|C|} \sum_{c \in C} {l^k_c})^2 \\
&&\sum_{i \in \mathcal{V}} x_{i} > \beta,
\end{eqnarray}
where $B$ is the batch size and $x_i$ is the number of training batches for node $i$. In this study, we assume that the node receiving the model randomly samples its non-IID data to construct mini-batches of size $B$, which are then used for training in that round. Consequently, $\frac{Bx_{i}}{N_{i}} L_{i}$ reflects the expected number of samples labeled $c$ that will be utilized by node $i$ for training the model. 
Therefore, by determining the node according to \eqref{eq:opt}, the expected value of the variance of $L^{k+1}$ can be minimized. When the variance becomes 0, our objective \eqref{eq:obj} is fulfilled. That is, the distribution of $L^{k+1}$ will be a uniform distribution, which suggests that all labels have been used equally for model training up to cycle $k+1$. This uniform distribution is desirable as it implies unbiased learning based on all available labels, leading to a model with balanced performance across all classes. $\beta$ denotes the minimum aggregate number of training batches, and a constraint is imposed such that the cumulative count of training batches per cycle surpasses $\beta$, thereby averting the cessation of the training process. Note that since there are no constraints on the number of training batches for each node, scenarios may arise where $x_i=0$, indicating nodes that do not participate in training.

We remark that the allocation naturally accommodates extreme label skew. If node $i$ holds no samples of class $c$ ($l_{i,c}=0$), the expected contribution $\frac{Bx_{i}}{N_{i}} L_{i}$ simply contains no mass on class $c$, and the optimizer compensates by allocating more batches to the nodes that do hold class $c$. In the extreme case where including a node only increases the variance objective, the solution sets $x_i=0$ and the node is skipped in that cycle, while the constraint $\sum_{i} x_{i} > \beta$ guarantees that training never stalls. Our experimental setting, in which each node holds only 2--5 of the 10 labels, already covers the case where every node entirely lacks half or more of the classes.

The main objective of this paper is to demonstrate the effectiveness of the model circulation-based FL in improving the trade-off between model accuracy and communication overhead. Therefore, while efficient solutions for the optimization problem are not the focus of this study, we discuss here that the problem is manageable within reasonable constraints.
Since this optimization problem is an integer programming problem due to the integer constraint on \( x_i \), it is classified as NP-hard. This makes it challenging to obtain an exact solution when the number of nodes or the amount of training data per node is large. However, obtaining an exact solution is not essential for this optimization problem. If there is a significant class imbalance in the training data, it could impact model performance, but otherwise, the effect is expected to be minimal. Thus, it is sufficient to design a heuristic that provides an approximate solution, even with a loose approximation ratio.

Moreover, the main application scenario of this circulation-based distributed FL is cross-silo FL, where the number of participants is typically limited to a few dozen, and the dataset sizes are relatively small. (If nodes possessed large datasets, they would likely train models individually without participating in FL.) Therefore, exploratory metaheuristics such as random search are also valid options. Specifically, if the value of \( \sigma(L^{k}) \) is close to zero in cycle 0, exploring the neighborhood of the solution obtained in cycle 0 for subsequent cycles \( k \geq 1 \) may lead to efficient discovery of good solutions. In the experimental evaluation of this study, since both the number of nodes and the data volume per node are small, we determine the mini-batch size through exhaustive search.

\subsection{Route Determination}\label{ss:model_routing_alg}
In the route determination step, the circulating path of the model is determined for nodes to which the number of batches has been allocated in the mini-batch number allocation step. Considering that the data on clients is non-IID, following a statically determined route without accounting for data distribution can lead to a model biased towards certain classes of data in the short term, significantly deviating from the global optimum. To address this, our proposed methodology optimizes the circulating route of the model. This prevents the model from being updated in an imbalanced manner across different classes during its circulation among nodes, thereby averting short-term biases in the model.

As a premise for determining the routing path, nodes with fewer training batches tend to exhibit a greater bias in data distribution (stronger non-IID). This is because our proposed aim is to minimize the bias in the label distribution of data used for training. Consequently, nodes with less bias in their data label distribution end up having a higher number of training batches. Conversely, nodes with a more significant bias in their data label distribution have fewer training batches. Therefore, training nodes with fewer batches early in the cycle is anticipated to skew the data distribution used for model updates in the short term. To counteract this, we propose a routing path that trains nodes with a higher number of batches first, thus avoiding short-term training data bias.

Furthermore, this route determination is expected to prevent catastrophic forgetting. Our proposed Tram-FL learns by circulating a single model throughout the network. Therefore, once a node has trained the model, its data is not used for training again until the model returns to that node. This interim period could lead to catastrophic forgetting. The training sequence determined by the route determination implies that nodes trained earlier are more likely to receive and train the model later, increasing the risk of catastrophic forgetting. However, by prioritizing training at nodes with larger batch numbers, our route determination potentially reduces this risk, as nodes with higher catastrophic forgetting risk undergo more extensive batch training.

We impose a constraint in our route determination where the last node of a cycle continues training consecutively. Specifically, if node 'i' is the last to train in cycle t-1, it is also the first to train in cycle t. This approach is expected to reduce communication frequency by one cycle per round, leading to a maximum communication cost reduction of 1/n, considering 'n' as the total number of nodes. Given our focus on a cross-silo setting with fewer nodes, this constraint is anticipated to reduce communication traffic significantly.

Our route determination assumes that any node can forward the model to any other node, which holds for cross-silo FL over the Internet, where connectivity is logical rather than physical. If the overlay is sparse, the route must be chosen from the feasible successors only, and a strongly constrained topology (e.g., a line graph) may force label-imbalanced orderings, degrading the short-term balance that our route rule provides. Because Steps~1 and~2 are executed at the beginning of every cycle, the formulation permits per-cycle re-optimization under the current connectivity, which is a natural basis for dynamic route adaptation.

Finally, we note the scope of the route determination in this paper. It adopts a deterministic rule that orders nodes by their allocated batch counts, which suppresses short-term label bias and mitigates the forgetting risk as described above. A full treatment of route optimization, however, should jointly account for per-node resource constraints such as heterogeneous computation capabilities, time-varying communication bandwidths, and node availability, all of which interact with the batch allocation and constitute a training-scheduling problem in its own right. We therefore leave the joint optimization of the circulation route under resource constraints to future work. Indeed, our follow-up study on load-aware training scheduling for model-circulation-based DFL~\cite{kainuma2025load} addresses a part of this problem by scheduling training under computation and communication load variations, and experimentally confirms that the node selection and ordering strategy materially affects convergence.

\subsection{Momentum Compression}\label{ss:momentum_comp}
In Tram-FL, momentum is computed and transmitted to the subsequent node during training. This approach is predicated on the findings indicating that incorporating the concept of momentum into FL, as demonstrated in \cite{karimireddy2020scaffold, balu2021decentralized, yu2019linear}, enhances model accuracy. However, direct transmission of momentum increases communication costs, thereby creating a trade-off between model accuracy and communication overhead. Consequently, we adopted a methodology wherein the momentum is compressed prior to transmission based on the premise that the precision of momentum does not significantly impact training.

For momentum compression, we employed quantization, a technique commonly used for compressing DNN models. Quantization reduces the bit-width of model parameters, which are typically represented as 32-bit floating-point numbers, to 8-bit or even lower integer values. Various quantization methods for DNNs have been proposed \cite{Wu2020IntegerQF}. We adopt dynamic affine quantization because it is stateless, incurs negligible computational overhead, and bounds the element-wise approximation error, which suits the momentum whose precision requirement is loose. Alternative compression schemes are complementary rather than competing: sparsification (e.g., top-$k$) can achieve higher compression ratios but requires error-feedback memory at each node, which conflicts with our design in which nodes are stateless between visits; knowledge-distillation-based compression transfers knowledge rather than parameters and presupposes an auxiliary dataset at each node. Combining such techniques with model circulation is left as future work, since they apply equally to all DFL baselines and do not affect the relative comparison in this paper.

The conversion of a real value $v \in [p, q]$ into a signed $b$-bit integer value $v_q \in \{-2^{b-1}, -2^{b-1}+1, \ldots, 2^{b-1}-1\}$ can be expressed using the following formula:
\begin{eqnarray}
v_{q} &=& \mathrm{clip} ( \mathrm{round}(s \cdot v + z) , -2^{b-1}, 2^{b-1}, -1)\\
\mathrm{clip}(v, l, u) &=& \left \{ \begin{array}{l}
l, ~ v<l \\
v, ~ l \leq v \leq u \\
u, ~ v>u
\end{array} \right. \\
s &=& \frac{2^{b}-1}{q-p} \\
z &=& -\mathrm{round}(p\cdot s)-2^{b-1}.
\end{eqnarray}
Here, $s$ represents the scale factor, and $z$ denotes the zero-point, which is determined by the range $[p, q]$ of the quantization target, as well as by the bit length after quantization. "round" means to convert a real number to the nearest integer value. Conversely, the formula for dequantization is represented as follows:
\begin{eqnarray}
\hat{v} &=& \frac{1}{s}(v_{q}-z).
\end{eqnarray}
It is important to note that obtaining an exact replica of the original input $v$ through dequantization is challenging due to the rounding of values during quantization. Therefore, the above formula yields only an approximate value $\hat{v} \approx v$.

In a manner akin to momentum, models can also be quantized, a known technique for reducing communication costs during model transmission. Previous research indicates that even when models are quantized to 8 bits, the accuracy degradation is limited to approximately 1\%. Consequently, applying model quantization to the proposed methodology is likely to yield a reduction in communication costs with minimal impact on model accuracy. However, since model quantization is applicable to most existing DFL methods, applying it to our proposed approach could lead to a lack of uniqueness and compromise fairness in comparative analysis. Consequently, model compression is not considered within the scope of this paper.

\subsection{Summary}\label{ss:summary}
In summary, each cycle of Tram-FL proceeds as follows. The node initiating the cycle solves the mini-batch number allocation to assign the number of training batches $x_i$ to every node (Step~1) and orders the nodes with allocated batches in descending order of $x_i$, keeping the last node of the previous cycle first (Step~2). The model and the momentum then circulate along this route: each node performs its allocated momentum-SGD updates (Step~3) and forwards the model together with the quantized momentum to the next node (Step~4). The cycles repeat until the validation accuracy reaches the target or the communication budget is exhausted.

\section{EVALUATION}\label{s:evaluation}
\subsection{Experimental setup}
In this study, we benchmarked Tram-FL against baseline methodologies, including PDMM-SGD, Gossip-SGD, and Centralized machine learning. We adopted image classification tasks using the MNIST \cite{lecun1998gradient} and CIFAR-10 \cite{krizhevsky2009learning} datasets. MNIST comprises images of handwritten digits across 10 classes, while CIFAR-10 consists of color images of objects belonging to 10 classes. The sample sizes for the training and test sets were {60,000, 10,000} for MNIST and {50,000, 10,000} for CIFAR-10, respectively.  We emulated a mesh network with five nodes, randomly assigning 2 to 5 data labels to each node, ensuring that all labels were allocated across the nodes. Samples were distributed to each node according to the label allocation, with each node receiving all samples associated with its distributed data labels. Consequently, the number of samples increased proportionally to the number of assigned labels.

\begin{table}
  \caption{DNN structure for MNIST}
  \label{tb:DNN_MNIST}
  \centering
  \begin{tabular}{l}
  \hline
    32 ch Conv ($3\times3$, valid), ReLU \\ \hline
    64 ch Conv ($3\times3$, valid), ReLU \\ \hline
    64 ch Max pooling ($2\times2$), Dropout (0.25), Flatten \\ \hline
    Dense (128 neurons), ReLU, Dropout (0.5), Dense (10 neurons), Softmax \\ \hline
  \end{tabular}
\end{table}

\begin{table}
  \caption{DNN structure for CIFAR-10}
  \label{tb:DNN_cifar10}
  \centering
  \begin{tabular}{l}
  \hline
    32 ch, Conv ($3\times3$, same size), Group Norm (32 ch $\times$ 1group), ReLU \\ \hline
    64 ch, Conv ($3\times3$, same size), Group Norm (32 ch $\times$ 2groups), ReLU \\ \hline
    64 ch, Max pooling ($2\times2$), Dropout (0.25) \\ \hline
    64 ch, Conv ($3\times3$, same size), Group Norm (32 ch $\times$ 2groups), ReLU \\ \hline
    64 ch, Conv ($3\times3$, same size), Group Norm (32 ch $\times$ 2groups), ReLU \\ \hline
    64 ch, Max pooling ($2\times2$), Dropout (0.25), Flatten \\ \hline
    Dense (512 neurons), ReLU, Dropout (0.5), Dense (10 neurons), Softmax \\ \hline
  \end{tabular}
\end{table}

\begin{table}
  \caption{Experimental setup}
  \label{tb:experimental_setup}
  \centering
  \begin{tabular}{l|ll}
  \hline
    Dataset & MNIST & CIFAR-10 \\ \hline
    Number of nodes & 3, 5, 10 & 3, 5, 10 \\ \hline
    Model size & 1,199,882 & 2,168,746 \\ \hline
    Input size & 784 & 1024 \\ \hline
    Mini-batch size & 100 & 100 \\ \hline
    Learning-rate & 0.005 & 0.005 \\ \hline
    $\alpha$ & 0.9 & 0.9 \\ \hline
    $\beta$ & 15, 25, 50 & 15, 25, 50 \\ \hline
  \end{tabular}
\end{table}

The DNN architectures employed for training on the MNIST and CIFAR-10 datasets are presented in \cref{tb:DNN_MNIST} and \cref{tb:DNN_cifar10}. In such scenarios, batch normalization \cite{ioffe2015batch} is customarily inserted before each convolution layer as the de facto standard. However, when utilizing non-IID datasets, the statistics of mini-batch sets are skewed, making group normalization \cite{wu2018group}, layer normalization \cite{ba2016layer}, and instance normalization \cite{ulyanov2016instance} more appropriate choices. For the CIFAR-10 DNN structure, we employed group normalization in our normalization process, which utilizes the mean/variance of channel groups. The initial weights, $w^{(0)}$, were initialized using He's method \cite{he2015delving}. In PDMM-SGD, a common seed value was utilized for initialization across all nodes. Momentum was compressed using 8-bit quantization. Refer to \cref{tb:experimental_setup} for additional parameters.

Furthermore, we investigated the relationship between communication cost and model accuracy due to momentum compression. This analysis was conducted using a 5-node mesh network emulated with CIFAR-10. Momentum was originally 32-bit and was compressed to 16-bit, 8-bit, 4-bit, and 2-bit through quantization. The other settings remained consistent with those previously described.

\subsection{Experimental result}
\subsubsection{Performance comparison with benchmarks}
\begin{figure*}[!t]
    \centering
    \begin{tabular}{ccc}
    \subfloat[MNIST]{\includegraphics[width=0.72\columnwidth]{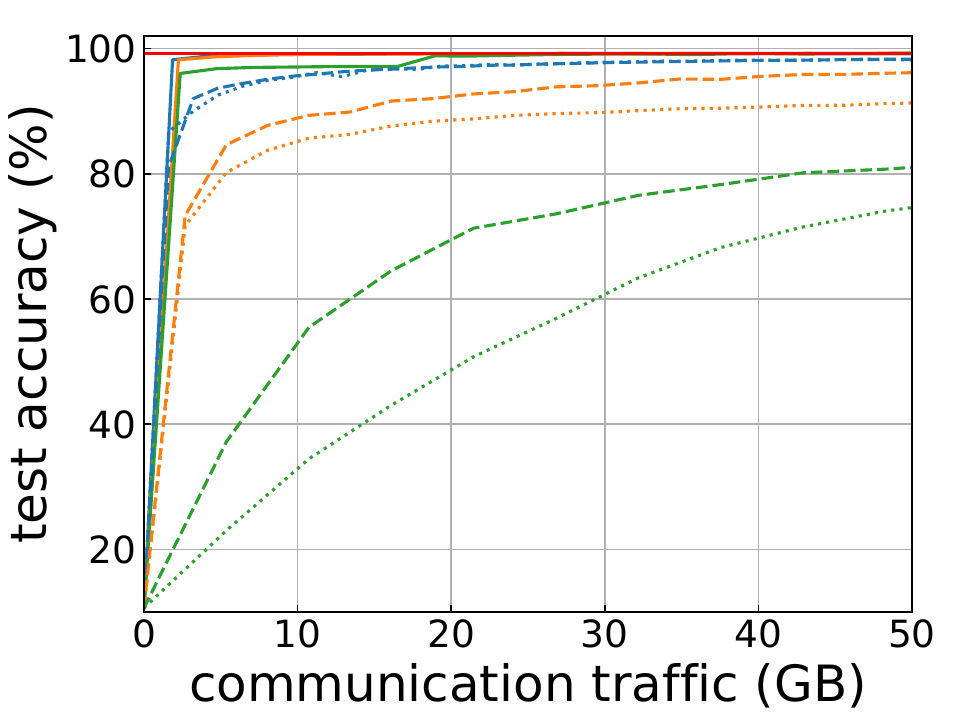}} &
    \subfloat[CIFAR-10]{\includegraphics[width=0.72\columnwidth]{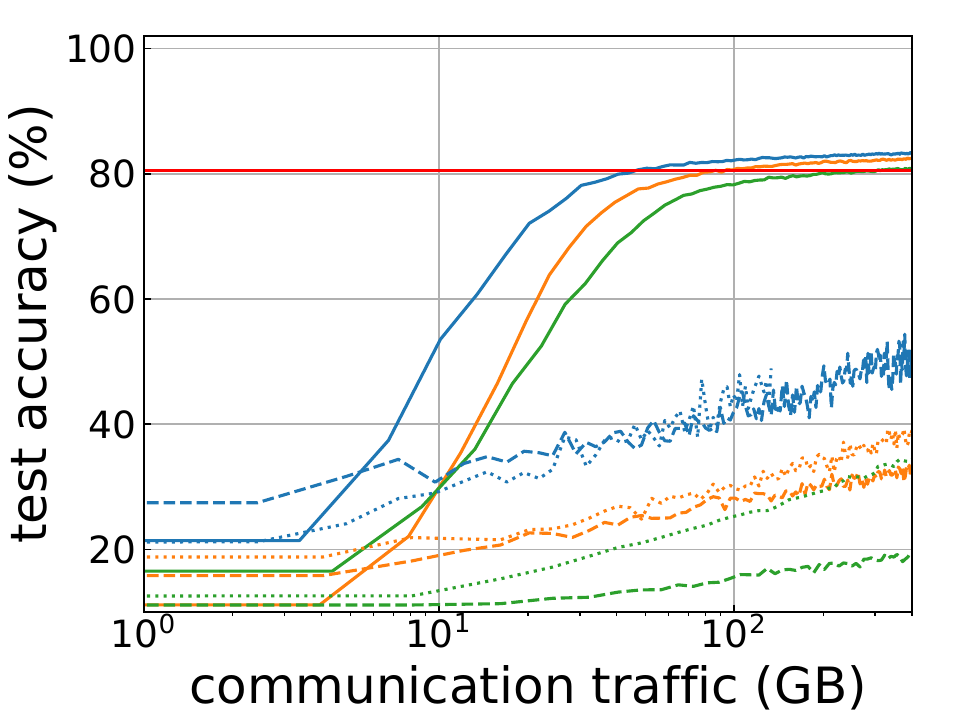}} &
    \subfloat{\includegraphics[width=0.45\columnwidth]{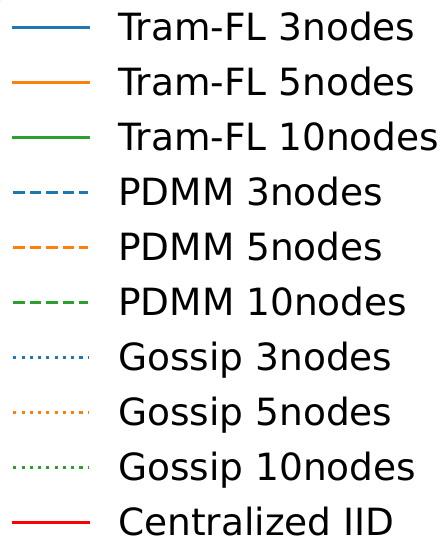}} \\
    \end{tabular}
    \caption{Test accuracy as a function of the communication traffic, utilizing non-IID data sets. Evaluation includes (a) MNIST for handwritten digit classification and (b) CIFAR-10 for object recognition. 8-bit quantization was used for momentum compression.}
    \label{fig:acc_comm}
\end{figure*}

\begin{figure*}[!t]
    \centering
    \begin{tabular}{ccc}
    \subfloat[MNIST]{\includegraphics[width=0.72\columnwidth]{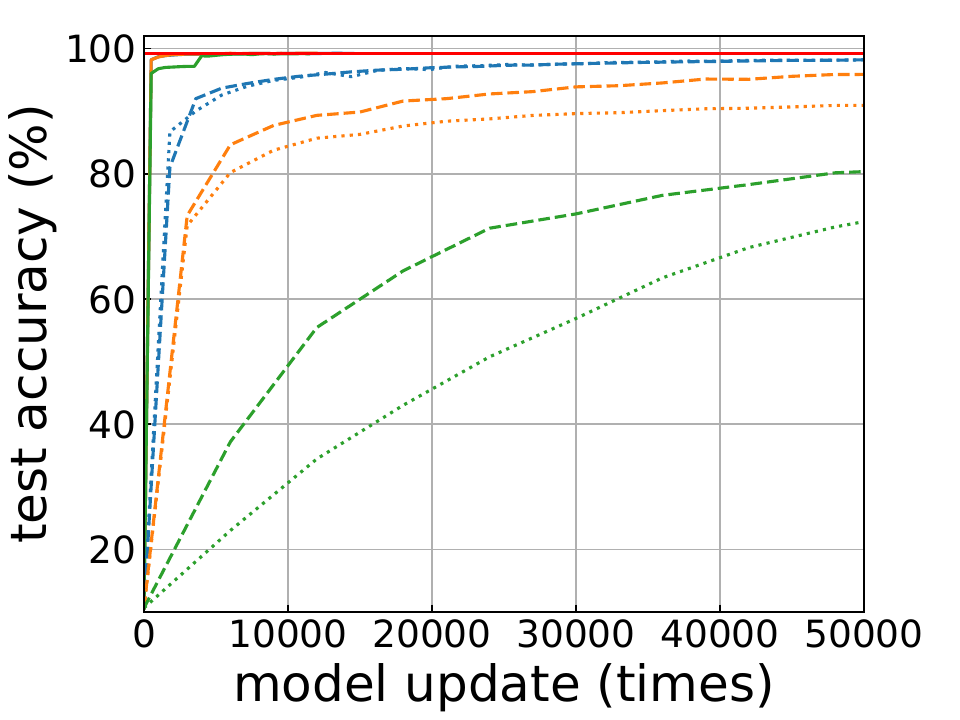}} &
    \subfloat[CIFAR-10]{\includegraphics[width=0.72\columnwidth]{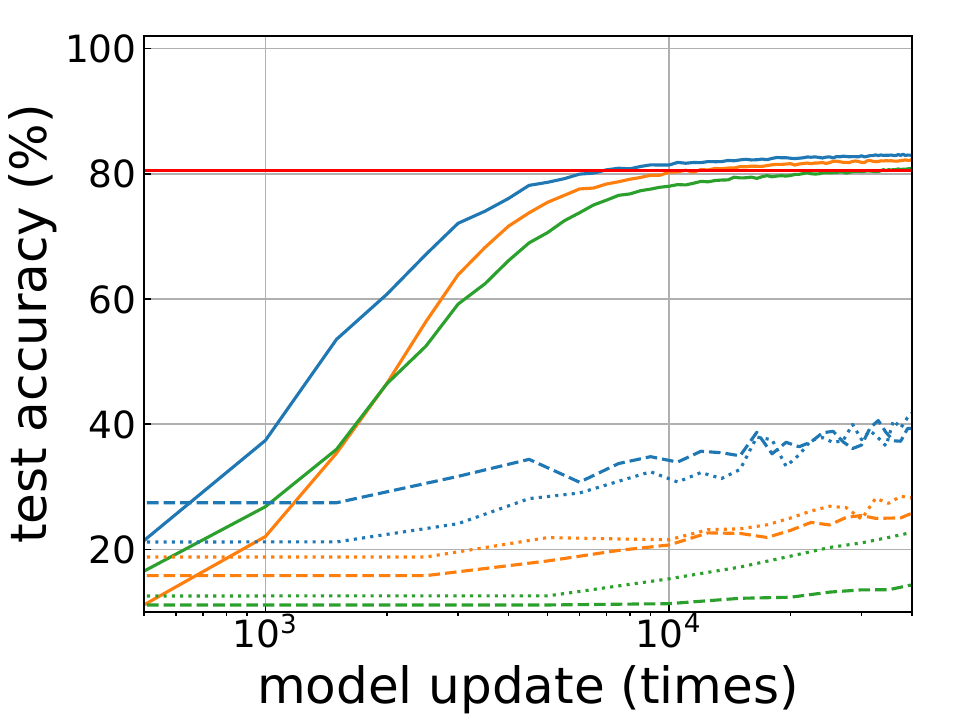}} &
    \subfloat{\includegraphics[width=0.45\columnwidth]{figure/legend.pdf}} \\
    \end{tabular}
    \caption{Test accuracy as a function of the total number of model updates, utilizing non-IID data sets. Evaluation includes (a) MNIST for handwritten digit classification and (b) CIFAR-10 for object recognition. 8-bit quantization was used for momentum compression.}
    \label{fig:acc_update}
\end{figure*}

\cref{fig:acc_comm} illustrates the test accuracy relative to communication traffic. In the Gossip-SGD and PDMM-SGD algorithms, the accuracy of the global model was determined by averaging the accuracies of all models within the network. As communication traffic escalates, the model undergoes updates and advances in learning, culminating in enhanced test accuracy across all tasks. The experimental outcomes for MNIST and CIFAR-10 are respectively delineated in \cref{fig:acc_comm} (a) and (b). The centralized IID (red solid line) represents the accuracy achieved when all datasets are consolidated at a single location for training. In every setting, Tram-FL converged with less communication traffic compared to Gossip-SGD and PDMM-SGD. An increase in the number of nodes tends to deteriorate overall convergence. It is particularly noteworthy that in Gossip-SGD and PDMM-SGD, the impact of varying node counts is pronounced, significantly diminishing accuracy improvement when the node count reaches 10. Conversely, Tram-FL exhibits relatively minor effects from changes in node count, consistently achieving the accuracy of centralized IID even with 10 nodes.

It may appear counterintuitive that a decentralized method matches, and in some runs slightly exceeds, the centralized-IID reference. Two factors explain this behavior. First, the mini-batch number allocation makes the label usage exactly balanced over cycles, whereas centralized training samples mini-batches uniformly from the pooled dataset, whose label proportions fluctuate batch by batch; balanced label exposure can yield marginally better generalization on the class-balanced test set. Second, the restricted number of updates per cycle with circulated momentum acts as an implicit regularization, reducing overfitting to any particular subset of the data. We emphasize that the observed differences are marginal and should be interpreted as Tram-FL attaining centralized-level accuracy rather than as evidence of systematic superiority over centralized training.

\cref{fig:acc_update} demonstrates the test accuracy in relation to the total number of model updates. \cref{fig:acc_update} (a) and (b) respectively present the experimental results for MNIST and CIFAR-10. In every scenario, Tram-FL achieved convergence with fewer model updates than Gossip-SGD and PDMM-SGD, signifying a reduction in computational costs. Additionally, akin to the results in \cref{fig:acc_comm}, Tram-FL demonstrated minimal impact from node count variations and exhibited stable convergence characteristics.

\cref{tb: result} presents the communication traffic and the number of updates required to achieve a specified level of accuracy. For the MNIST dataset, thresholds were set at 99\%, 90\%, and 80\%, while for the CIFAR-10 dataset, thresholds were established at 80\%, 60\%, and 40\%. In the Gossip-SGD and PDMM-SGD algorithms, the accuracy of the global model was determined by averaging the accuracies of all models within the network. The "-" category represents scenarios in which the thresholds were not attained. In the experiment utilizing the MNIST dataset with Tram-FL, the interval for testing was set after every 500 model updates. It is noteworthy that the thresholds of 80\% and 90\% were achieved prior to the initial 500 model updates. It was observed that the Gossip-SGD and PDMM-SGD algorithms failed to surpass certain thresholds in some scenarios. In contrast, Tram-FL exceeded all the designated thresholds with lower communication traffic and fewer updates. When comparing only scenarios where thresholds were exceeded with PDMM-SGD, the communication traffic for MNIST was reduced by an average of 84.7\%, and the number of model updates by 96.8\%. For CIFAR-10, the reductions were 83.6\% in communication traffic and 96.1\% in the number of model updates, respectively. Considering that scenarios not surpassing the thresholds would require additional communication and computation to do so, these figures represent minimum values, suggesting that actual communication and computation costs could be further reduced. These measured reductions are also consistent with the analytical comparison in \cref{tb:overhead_comp}: the savings stem from transmitting and training a single model rather than $n$ models in parallel.

\begin{table*}
  \caption{Communication traffic and number of updates to reach a certain accuracy.\\“-“ means the threshold was not attained.}
  \label{tb: result}
  \centering
  \begin{tabular}{lll||ccc|ccc|ccc}
  \hline
    &&& \multicolumn{3}{c|}{Tram-FL} & \multicolumn{3}{|c|}{PDMM-SGD} & \multicolumn{3}{|c}{Gossip-SGD} \\
    && Acc. & 3nodes & 5nodes & 10nodes & 3nodes & 5nodes & 10nodes & 3nodes & 5nodes & 10nodes \\ \hline
    \multirow{6}{*}{MNIST} & \multirow{3}{*}{Communication traffic (GB)} & 99\% & 9.31 & 9.03 & 18.19 & - & - & - & - & - & -  \\
    &  & 90\% & 1.86 & 2.24 & 2.36 & 3.20 & 16.09 & - & 3.22 & 32.18 & - \\
    &  & 80\% & 1.86 & 2.24 & 2.36 & 1.59 & 5.36 & 42.91 & 1.61 & 5.36 & - \\ \cline{2-12}
    & \multirow{3}{*}{Model update (times)} & 99\% & 2500 & 2000 & 3800 & - &  - & - & - & - & - \\
    &  & 90\% & 500 & 500 & 500 & 3600 & 18000 & - & 3600 & 36000 & - \\
    &  & 80\% & 500 & 500 & 500 & 1800 & 6000 & 48800 & 1800 & 6000 & - \\ \hline
    \multirow{6}{*}{CIFAR-10} & \multirow{3}{*}{Communication traffic (GB)} & 80\% & 43.09 & 75.69 & 184.04 & - & - & - & - & - & -  \\
    &  & 60\% & 14.14 & 23.68 & 31.29 & - & - & - & - & - & -  \\
    &  & 40\% & 8.75 & 15.05 & 15.98 & 53.32 & - & - & 58.17 & - & -  \\ \cline{2-12}
    & \multirow{3}{*}{Model update (times)} & 80\% & 6400 & 9600 & 20300 & - & - & - & - & - & -  \\
    &  & 60\% & 2100 & 3000 & 3500 & - & - & - & - & - & -  \\
    &  & 40\% & 1300 & 1900 & 1800 & 33000 & - & - & 36000 & - & -  \\ \hline
  \end{tabular}
\end{table*}

\subsubsection{Impact of Compression Rates on Model Accuracy}
\begin{figure}[!t]
\centering
\includegraphics[width=1\columnwidth]{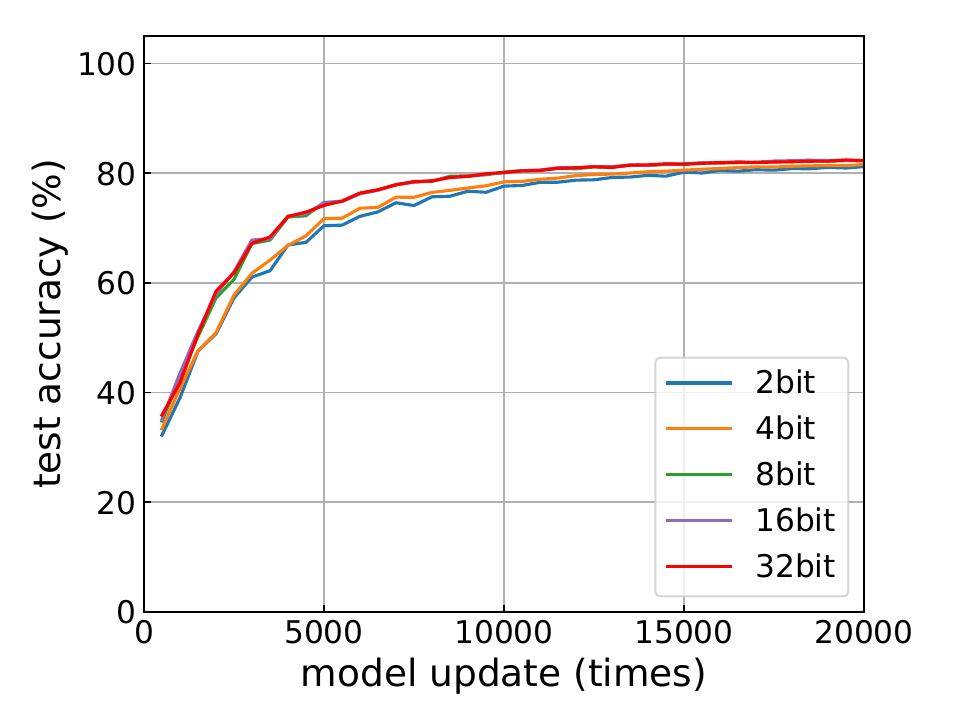}
\caption{Test accuracy as a function of the total number of model updates across different bits of model quantization. 5-node mesh network with CIFAR-10.}
\label{fig:quantization_convergence}
\end{figure}

\cref{fig:quantization_convergence} illustrates the relationship between the cumulative number of model updates within the network and the corresponding test accuracy for the initial scenario. This data represents the mean of 15 trials conducted with varying seed values. It is observed that as the total number of model updates between nodes increases, the model undergoes continuous updates and learning, resulting in enhanced test accuracy across all tasks. Remarkably, there is a negligible difference in the accuracy per model update among the 32-bit non-quantized momentum, 16-bit quantized momentum, and 8-bit quantized momentum. Furthermore, although the 4-bit and 2-bit quantized momentum requires a greater number of updates to converge, the final test accuracy is virtually identical to that of the non-quantized momentum. The communication traffic per transmission is significantly reduced by quantizing the momentum, while the model accuracy remains largely unaffected. Consequently, this approach holds promise for substantial reductions in overall communication costs.

\begin{figure}[!t]
\centering
\includegraphics[width=1\columnwidth]{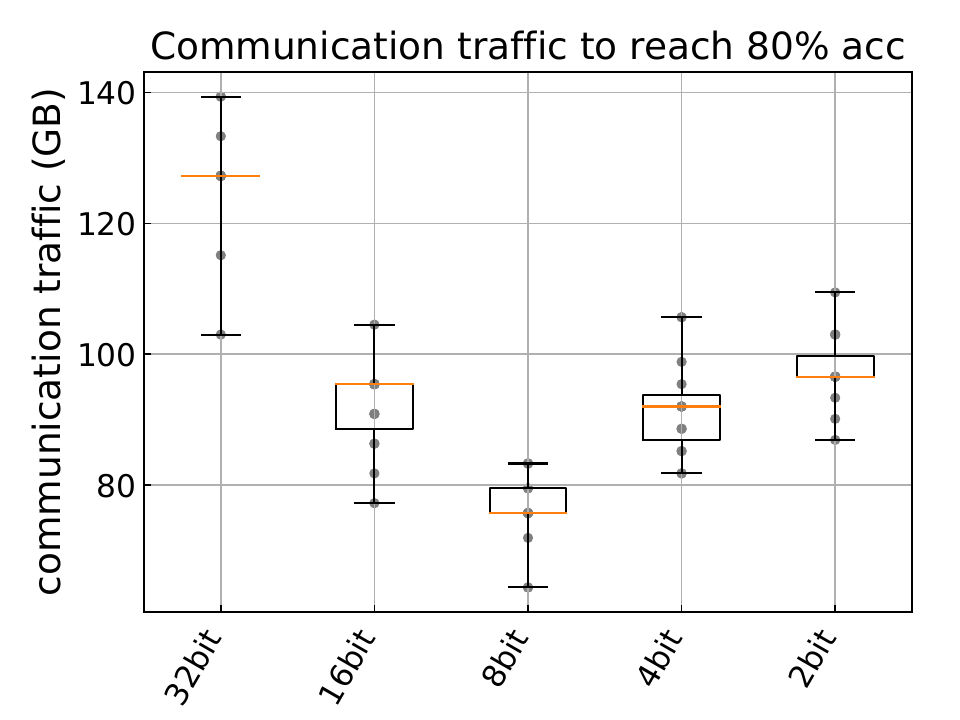}
\caption{Comparison of the number of model transmissions required to reach an 80\% accuracy level, illustrated across different bits of model quantization.}
\label{fig:quantization_hakohige}
\end{figure}

\cref{fig:quantization_hakohige} presents a box-and-whisker plot depicting the communication traffic (in GB) at the point where test accuracy reaches 80\%. This data is based on an average of 15 trials conducted with varying seed values. All methods employing momentum quantization demonstrated a reduction in communication traffic compared to the approach without momentum quantization. Notably, the 8-bit quantized momentum method achieved an average reduction of 40.09\% in communication traffic compared to the non-quantized approach.

\section{CONCLUSION}\label{s:conclusion}
In this paper, we propose a novel DFL methodology based on model circulation, termed Tram-FL, which incorporates the transmission of compressed momentum. We also present an example of the mini-batch number allocation and a compatible route determination.
Tram-FL operates by sharing a single model across the entire network. The learning process is advanced by circulating this model with an optimal number of training batches and a designated circulation route, thereby realizing a global model.
Our experiments demonstrate that while the compression of momentum predominantly contributes to a substantial reduction in communication traffic, it does not significantly compromise accuracy. Furthermore, Tram-FL successfully acquires an accurate global model with minimal communication and computational costs in various scenarios, even when non-IID datasets are distributed among nodes.
This suggests that efficient learning on non-IID datasets is feasible if an optimal mini-batch number allocation is established, enabling the selection of the ideal training batch size, and a route determination is implemented for choosing the most effective circulation route.
We conclude by summarizing the limitations of this work. First, Tram-FL targets cross-silo FL; the sequential circulation limits parallelism, and the wall-clock time of a cycle grows linearly with the number of nodes. Second, the mini-batch number allocation relies on pre-shared label histograms, whose privacy implications were discussed but not quantitatively evaluated. Third, the evaluation assumes reliable links and no node churn. Fourth, the route determination adopts a deterministic rule; its joint optimization with per-node resource constraints (heterogeneous computation capabilities, time-varying bandwidths, and node availability) is a training-scheduling problem in its own right, a part of which is addressed by our follow-up work~\cite{kainuma2025load}.

Our future research agenda is oriented towards adapting the proposed method to more realistic scenarios. Specifically, we aim to establish methods for sharing label distributions in scenarios where data at each node increases during training, and techniques to ensure that all nodes contribute sufficiently to the training process.

\bibliographystyle{IEEEtran}
\bibliography{main}

\end{document}